\documentclass{article}

\usepackage[dblblindworkshop, final]{neurips_2026}

\usepackage[utf8]{inputenc} %
\usepackage[T1]{fontenc}    %
\usepackage{hyperref}       %
\usepackage{url}            %
\usepackage{booktabs}       %
\usepackage{amsfonts}       %
\usepackage{amsmath}
\usepackage{nicefrac}       %
\usepackage{microtype}      %
\usepackage[table]{xcolor}
\usepackage{multirow}
\usepackage{graphicx}

\newcommand{\methname}{\textsc{IrekoGPT}}

\usepackage{cleveref}

\crefname{section}{Sec.}{Secs.}
\Crefname{section}{Sec.}{Secs.}

\crefname{subsection}{Sec.}{Secs.}
\Crefname{subsection}{Sec.}{Secs.}

\crefname{figure}{Fig.}{Figs.}
\Crefname{figure}{Fig.}{Figs.}

\crefname{table}{Tab.}{Tabs.}
\Crefname{table}{Tab.}{Tabs.}

\crefname{equation}{Eq.}{Eqs.}
\Crefname{equation}{Eq.}{Eqs.}

\crefname{algorithm}{Alg.}{Algs.}
\Crefname{algorithm}{Alg.}{Algs.}

\crefname{appendix}{App.}{Apps.}
\Crefname{appendix}{App.}{Apps.}

\title{\methname: Turning Structured Pruning into Post-Hoc Slimmable LLMs}
\workshoptitle{AXIOM: Foundations of Efficient Deep Learning}

\author{%
Pietro Moriello \quad Pietro Buzzega \quad Angelo Porrello \quad Simone Calderara \\[8pt]
AImageLab, University of Modena and Reggio Emilia, Italy \\
\texttt{\{name.surname\}@unimore.it}
}

\begin{document}

\maketitle

\begin{abstract}
We introduce \methname{}, a post-hoc method for converting pretrained LLMs into slimmable models whose width can be adjusted at inference time. Building on SliceGPT, we retain its projection matrices without pruning them, allowing a single model to expose nested subnetworks at different widths. We improve robustness by calibrating each layer across multiple compression ratios, and correct downstream linear layers through gradient-free ridge regression. Across Llama and Qwen models, preliminary results show improvements over naive PCA-based slimming, with the largest gains at high compression. 

Code is available at \url{https://github.com/aimagelab/IrekoGPT}
\end{abstract}

\section{Introduction}
\label{sec:intro}
Deploying large neural networks on resource-constrained devices, such as mobile phones or embedded systems, requires effective compression techniques -- such as weight quantization~\cite{frantar2023gptq} and network pruning~\cite{han2016deepcompression} -- to enable efficient inference and democratize access to AI. Among these, \textbf{structured pruning}~\cite{ma2023llmpruner,ashkboos2024slicegpt} enables substantial savings without relying on dedicated hardware support, as compressed models can be obtained through simple operations in weight space, such as slicing the original weight matrices. Some structured approaches merge redundant neurons with similar structure~\cite{ma2023llmpruner,verma2025dotresize}, while others leverage eigenvector decompositions to identify the least important components to remove~\cite{wang2025svdllm,ashkboos2024slicegpt}. 

The resulting networks retain most of their original functionality while being lighter to store and faster. However, these approaches typically optimize the pruned architecture for a \textbf{predefined target size} (e.g., 80\% of the original model). In real-world scenarios, resource availability is rarely static or known a priori: it may fluctuate, for instance, depending on the device's current power-saving mode or on the workload it is required to handle at a given time. In these cases, structured pruning would require maintaining a separate compressed network for each resource budget, incurring additional storage costs and runtime overhead from loading and unloading different models. To overcome these limitations, pioneering works on \textbf{slimmable neural networks}~\cite{yu2019slimmable,yu2019universallyslimmable} train a single model to operate at multiple widths. However, this capability is typically learned from scratch, which conflicts with the growing reliance on pretrained foundation models.

In this paper, we address the problem of converting pretrained large language models into slimmable variants, as recently explored in~\cite{wong2026a3,zaccone2026flexrank,rauba2026nsn}. In particular, we revisit the established structured pruning method SliceGPT~\cite{ashkboos2024slicegpt} and show that its formulation can be naturally extended to support slimmable inference. Building on this connection, we introduce a set of modifications that better align the activations of smaller subnetworks with those of the original full-width model. We show that the resulting approach yields particularly pronounced gains at higher compression levels.

\section{Background}
\label{sec:background}
Let $Z^\ell \in \mathbb{R}^{N \times L \times D}$ denote the activations at layer $\ell$ of a Transformer model, where $N$ is the number of examples, $L$ the sequence length, and $D$ the token dimension. SliceGPT reduces $D$ through a sequence of transformations: specifically, given a calibration dataset $\mathcal{X} \in \mathbb{R}^{N \times L}$ of text examples, SliceGPT extracts the activations at layer $\ell$ and performs PCA to compute the eigenvector matrix $Q^\ell \in \mathbb{R}^{D \times D}$, whose columns are sorted by decreasing eigenvalue. By retaining only the first $d$ columns of $Q^\ell$, we obtain a projection $Q^\ell_d \in \mathbb{R}^{D \times d}$ that yields the compressed representation $Z^\ell_d = Z^\ell Q^\ell_d \in \mathbb{R}^{N \times L \times d}$ while minimizing the reconstruction error
$\|Z^\ell - Z^\ell_d(Q^\ell_d)^T\|_F^2$:
\begin{equation}
Q^{\ell}_d =
\operatorname*{arg\,min}_{U_d}
\|Z^\ell - Z^\ell U_dU_d^T\|_F^2
\quad
\textrm{s.t.}
\quad U_d^TU_d = I_d,\; d < D.
\label{orthonormal-opt}
\end{equation}
We refer the reader to \cite{ashkboos2024slicegpt} for a detailed description of how these projections are absorbed into the model weights and subsequently sliced. For simplicity, throughout this work we identify slicing with retaining the first $d$ columns of each $Q^\ell$ matrix.

\noindent\textbf{Sequential approach.} The procedure above is applied \textbf{sequentially}, from the first layer to the last. Indeed, after computing PCA at a given layer, the resulting projection is absorbed into the model weights before moving to the next layer. Hence, the activation of a downstream layer $Z^{\ell'}$ is affected by the pruning applied at all preceding layers $\ell < \ell'$. In other words, the principal components computed at each step reflect the \emph{distributional shifts} induced by earlier pruning operations.

\noindent\textbf{Unaffected layers.} We remark that this procedure is not applied to every layer in the model, but only to either the input or output dimension of linear layers adjacent to LayerNorm. All remaining dimensions are left unchanged and continue to operate at full width.

\section{Method}
\label{sec:method}
\subsection{Revisiting SliceGPT: from slicing to slimming}
\label{subsec:revisiting-slicegpt}
Our first observation is that SliceGPT can be turned into a slimmable model by absorbing the full eigenvector matrix $Q^\ell$ into the model weights (hence, without slicing). Indeed, since the columns of each $Q^\ell$ are ordered by decreasing eigenvalue, this induces a \textbf{nested structure}: retaining the first $d$ columns always preserves the most important principal directions. Different widths can therefore be obtained simply by truncating each projection to a different number of columns, while ensuring that smaller subnetworks are nested within larger ones and retain the highest-variance components.

This establishes a direct connection between SliceGPT and slimmable networks: the ordered PCA basis used for structured pruning can instead be retained in full and exploited to support dynamic width selection at inference time. This observation forms the core foundation of our approach.
\subsection{Improving slimmability by expanding the calibration dataset}
\label{subsec:improving-slimmability}
To improve slimmability, we revisit the sequential nature of SliceGPT described above. In the original approach, once the transformation at layer $\ell$ has been computed, its output is propagated to the next layer and therefore determines the activations used to construct the calibration set for layer $\ell+1$. These activations define the PCA basis and ultimately the compression subspace of the next layer.

In a slimmable model, however, the output of layer $\ell$ is not fixed at inference time: it can be dynamically reduced according to different width ratios. This raises a fundamental question: \emph{which width should be propagated when calibrating the next layer?} If calibration is performed only at full width, while inference later operates at, e.g., $60\%$ width, the downstream layers are optimized on activations drawn from a different distribution than the one they will actually receive at inference time. We therefore expect a shift in the activations used to compute PCA at subsequent layers.

We introduce a simple modification to ensure that PCA captures the distributional shifts induced when the network operates at \emph{multiple} width configurations. The key idea is to \emph{expand} the calibration set with activations collected at different widths. In particular, we define a set of compression ratios $R = \{r_1, \dots, r_K\}\subset (0,1)$. Given the original calibration dataset $\mathcal{X}$, we maintain $K+1$ parallel activation streams, one for each compression ratio in $\bar R = R \cup \{1\}$, where $1$ denotes the full-width model. After computing the PCA basis at layer $\ell-1$, we truncate it according to each ratio $r_k \in \bar R$ and propagate the corresponding stream independently to layer $\ell$. Thus, the $k$-th stream consistently follows the same compression ratio throughout the network:
\begin{equation}
    Q^{n}_{d_k} = Q^{n}[:, :d_k],
    \qquad
    d_k = \lfloor r_k D \rfloor,
    \qquad
    Z^\ell_k =
    \hat f_\ell\!\left(
        \mathcal{X};
        Q^1_{d_k},\dots,Q^{\ell-1}_{d_k}
    \right),
\end{equation}
where $\hat f_\ell(\cdot; Q^1_{d_k},\dots,Q^{\ell-1}_{d_k})$ denotes the network up to layer $\ell$ when operated at compression ratio $r_k$. The activations produced by all $K+1$ streams are then concatenated $Z^\ell =\operatorname{concat}\left(Z^\ell_1,\dots,Z^\ell_{K+1}\right) \in \mathbb{R}^{(K+1)NL \times D}.$ and jointly used to compute the PCA basis at layer $\ell$.
    
The resulting basis is therefore optimized over the activation distributions induced by all considered widths, rather than by the full-width network alone. We refer to this procedure as \emph{Expanded Calibration Dataset} (ECD). In practice, since compression proceeds iteratively, re-computing activations from the network input is unnecessary. We reduce computational cost by caching the activations collected at layer $\ell-1$ and propagating each stream only through the sub-network between the previous compression point and the current one.
\subsection{Leveraging full-width signals for slim subnetworks}
\label{sec:ridge}
As discussed in \cref{sec:background}, compression only affects a subset of components in the model. Several linear layers -- \textit{e.g.}, the value projection in self-attention -- retain their full output dimension even when their input is compressed; that is, their output width $d_{\text{out}}$ remains equal to the original hidden size. We exploit these width-preserving points to introduce a representation-level constraint. Specifically, we adjust the corresponding weight matrices so that, when fed with activations from compressed streams, their outputs remain close in MSE to those produced by the full-width model. In this way, full-width activations serve as reference signals for correcting representations of slim subnetworks.

In formal terms, let $I_k \in \mathbb{R}^{B \times d_k}$ denote the input activations produced by the $k$-th compressed stream, where $d_k$ is the corresponding compressed width, and let $O \in \mathbb{R}^{B \times d_{\text{out}}}$ denote the full-width target outputs, with $d_{\text{out}}$ equal to the original hidden size. We seek a residual correction $\Delta W$ to the original weight matrix $W$ such that the compressed streams reproduce the full-width outputs as closely as possible. To keep the correction controlled, we regularize its magnitude:
\begin{equation}
\Delta W^\star =
\operatorname*{arg\,min}_{\Delta W}
\sum_{k=1}^{K}
\left\|
O - I_k(W^T + \Delta W^T)
\right\|_F^2
+
\lambda \|\Delta W\|_F^2.
\label{eq:ridge-correction}
\end{equation}
The objective jointly accounts for all $K$ compression ratios in the expanded calibration set, while the regularization term encourages the corrected weights to remain close to the original $W$. Since \cref{eq:ridge-correction} is a least-square problem, it admits an efficient closed-form solution.

\section{Experiments}
\label{sec:experiments}
\newcommand{\gr}[1]{\textcolor{black!70}{#1}}   %

\begin{table}[t]
\centering
\caption{WikiText-2 perplexity (lower is better) across models and sparsity levels. $L$ denotes Llama and $Q$ Qwen; subscripts indicate version and size. SliceGPT\textsuperscript{\textdagger} is not slimmable and requires a separate model for each sparsity level.
Bold indicates the best result among slimmable variants only.}
\label{tab:ppl-neurips-comparison}
\resizebox{0.99\textwidth}{!}{%
\begin{tabular}{
lrrrr
@{\hspace{1.2em}\hspace{1.2em}}
lrrrr
}
\toprule
& $L_{2,\mathrm{7B}}$
& $L_{3.1,\mathrm{8B}}$
& $L_{3.2,\mathrm{1B}}$
& $Q_{3,\mathrm{1.7B}}$
&
& $L_{2,\mathrm{7B}}$
& $L_{3.1,\mathrm{8B}}$
& $L_{3.2,\mathrm{1B}}$
& $Q_{3,\mathrm{1.7B}}$ \\
\midrule
\rowcolor{gray!15}
Sparsity & \multicolumn{4}{c}{\textbf{Dense}}
&
Sparsity & \multicolumn{4}{c}{\textbf{25\%}} \\
\midrule
\gr{Original}
& \gr{$5.47$} & \gr{$6.24$} & \gr{$9.75$} & \gr{$16.71$}
&
\gr{SliceGPT\textsuperscript{\textdagger}}
& \gr{$7.50$} & \gr{$13.73$} & \gr{$19.13$} & \gr{$21.53$} \\
PCA
& $\mathbf{5.47}$ & $6.25$ & $9.76$ & $16.74$
&
PCA
& $8.62$ & $18.69$ & $20.32$ & $24.43$ \\
ECD
& $5.47$ & $\mathbf{6.24}$ & $\mathbf{9.76}$ & $\mathbf{16.72}$
&
ECD
& $7.98$ & $15.09$ & $19.68$ & $\mathbf{22.76}$ \\
\methname{}
& $5.51$ & $6.41$ & $9.88$ & $17.03$
&
\methname{}
& $\mathbf{7.92}$ & $\mathbf{14.19}$ & $\mathbf{17.77}$ & $24.06$ \\
\midrule
\rowcolor{gray!15}
Sparsity & \multicolumn{4}{c}{\textbf{10\%}}
&
Sparsity & \multicolumn{4}{c}{\textbf{30\%}} \\
\midrule
\gr{SliceGPT\textsuperscript{\textdagger}}
& \gr{$5.93$} & \gr{$8.05$} & \gr{$12.69$} & \gr{$18.51$}
&
\gr{SliceGPT\textsuperscript{\textdagger}}
& \gr{$8.52$} & \gr{$19.41$} & \gr{$23.31$} & \gr{$23.39$} \\
PCA
& $6.26$ & $8.67$ & $13.26$ & $18.65$
&
PCA
& $10.06$ & $26.84$ & $24.99$ & $27.81$ \\
ECD
& $\mathbf{6.05}$ & $\mathbf{8.09}$ & $\mathbf{12.72}$ & $\mathbf{18.22}$
&
ECD
& $8.98$ & $20.18$ & $23.31$ & $\mathbf{25.12}$ \\
\methname{}
& $6.07$ & $8.27$ & $12.83$ & $18.82$
&
\methname{}
& $\mathbf{8.80}$ & $\mathbf{18.91}$ & $\mathbf{20.41}$ & $26.68$ \\
\midrule
\rowcolor{gray!15}
Sparsity & \multicolumn{4}{c}{\textbf{20\%}}
&
Sparsity & \multicolumn{4}{c}{\textbf{35\%}} \\
\midrule
\gr{SliceGPT\textsuperscript{\textdagger}}
& \gr{$6.82$} & \gr{$11.29$} & \gr{$16.36$} & \gr{$20.42$}
&
\gr{SliceGPT\textsuperscript{\textdagger}}
& \gr{$10.18$} & \gr{$27.02$} & \gr{$29.20$} & \gr{$26.38$} \\
PCA
& $7.63$ & $13.21$ & $17.43$ & $22.14$
&
PCA
& $12.20$ & $57.19$ & $32.27$ & $33.66$ \\
ECD
& $7.07$ & $11.68$ & $16.77$ & $\mathbf{20.83}$
&
ECD
& $11.01$ & $33.92$ & $29.91$ & $\mathbf{29.83}$ \\
\methname{}
& $\mathbf{7.06}$ & $\mathbf{11.64}$ & $\mathbf{15.89}$ & $21.81$
&
\methname{}
& $\mathbf{10.53}$ & $\mathbf{29.05}$ & $\mathbf{24.78}$ & $32.03$ \\
\bottomrule
\end{tabular}%
}
\end{table}

\textbf{Setup.} We use the original SliceGPT~\cite{ashkboos2024slicegpt} as our baseline and evaluate three additional variants: a PCA-only configuration, corresponding to SliceGPT without pruning; Expanded Calibration Dataset (ECD), which augments SliceGPT with the expanded calibration dataset introduced above; and our full method, \methname{}, which combines ECD with the ridge-based correction described in \cref{sec:ridge}. We evaluate four models: Llama-2-7B~\cite{touvron2023llama2}, Llama-3.2-1B~\cite{meta2024llama32}, Llama-3.1-8B~\cite{grattafiori2024llama3}, and Qwen3-1.7B~\cite{yang2025qwen3}. The calibration set consists of $128$ sequences of length $2048$ sampled from the WikiText-2 training split~\cite{merity2017wikitext}. For ECD, we use the same set of compression ratios across all models, $R=\{0.1,0.2,0.3\}$. For the ridge-based correction in \methname{}, we target the value projection $W_V$, which showed promising performance in preliminary experiments. 
To reduce VRAM usage, at each layer we randomly sample $B=65{,}536$ tokens ($25\%$ of the calibration tokens) for the regression step, and set $\lambda=1$ in all experiments. All experiments are run on cluster nodes equipped with NVIDIA A100 GPUs with $64$\,GB of VRAM.

\textbf{Evaluation.} Following SliceGPT~\cite{ashkboos2024slicegpt}, We evaluate all methods on language modeling by measuring perplexity on WikiText-2~\cite{merity2017wikitext}. We use a maximum sequence length of $2048$, truncating longer examples and concatenating shorter ones to match the target length.

\cref{tab:ppl-neurips-comparison} presents the main results. Notably, the $25\%$ and $35\%$ compression ratios are never included in the expanded calibration set, and therefore also assess generalization to unseen widths. We also report SliceGPT, although it is not slimmable and requires a separate compression procedure for each target sparsity. In contrast, the PCA-only variant is obtained by slimming a single checkpoint at inference time according to the procedure in \cref{subsec:revisiting-slicegpt}. Its performance remains reasonably close to SliceGPT, confirming that this construction already provides a viable slimmable model. Building on this baseline, ECD consistently improves performance and substantially reduces the gap with SliceGPT across models and sparsity levels, validating the benefit of calibrating over multiple widths. At higher compression, \methname{} further improves performance and, in several cases, matches or outperforms SliceGPT. We hypothesize that stronger compression induces larger shifts in the activation distribution, making both the expanded calibration and the subsequent correction increasingly beneficial. Qwen3-1.7B is the main exception, for which ECD alone remains preferable.

\begin{figure}[t]
    \centering
    \includegraphics[width=0.99\linewidth]{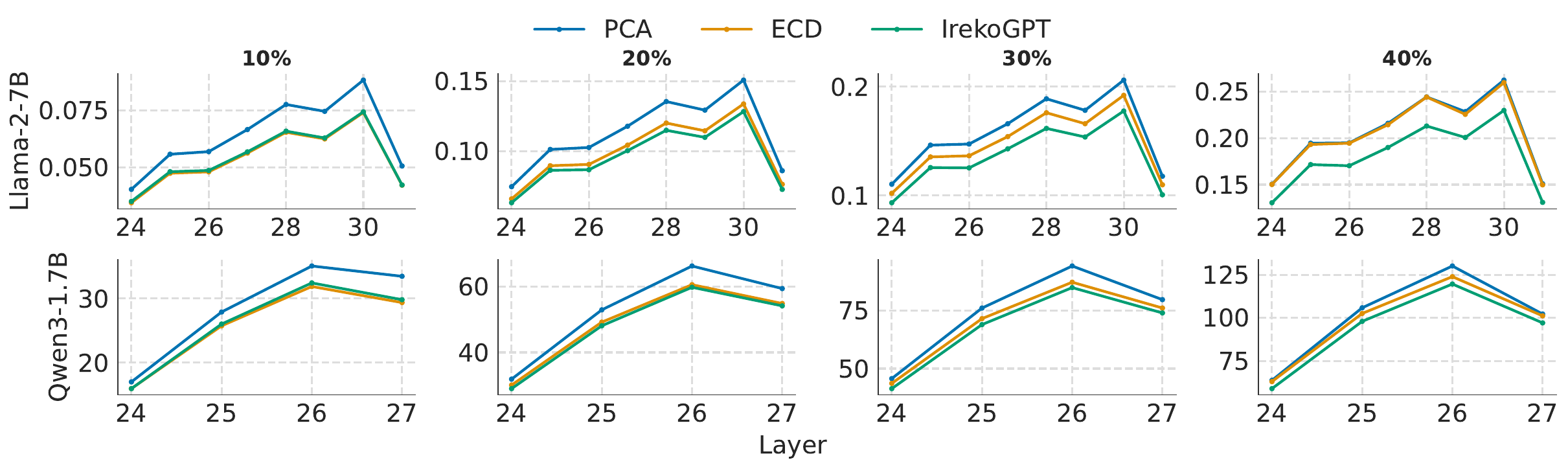}
    \vspace{-0.4em}
    \caption{Layer-wise MSE between compressed and full-width value-projection outputs across increasing sparsity levels. We compare PCA-only, ECD, and \methname{}; lower is better.}
\label{fig:mse_ireko}
    \vspace{-0.4em}
\end{figure}

\cref{fig:mse_ireko} provides a representation-level view of our method. Across sparsity levels and layers, \methname{} consistently reduces the MSE between compressed and full-width activations, often by a substantial margin over PCA-only and ECD. The gap becomes especially evident as sparsity increases, supporting our hypothesis that correcting compression-induced representation shifts is key to preserving the original model behavior.

\section{Conclusion}
\label{sec:conclusion}
We introduced \methname{}, a post-hoc approach for turning pretrained LLMs into slimmable models by reinterpreting SliceGPT's PCA structure. By calibrating across multiple widths and leveraging full-width signals, \methname{} improves robustness across compression levels while retaining a single dynamically configurable model.

\bibliographystyle{plainnat}
\bibliography{references}

\appendix
\section{Additional metrics plots.}

We present some additional plots to assess the behaviour of our method across a wider range of models and layers.

\begin{figure}[ht]
    \centering
    \includegraphics[width=0.99\linewidth]{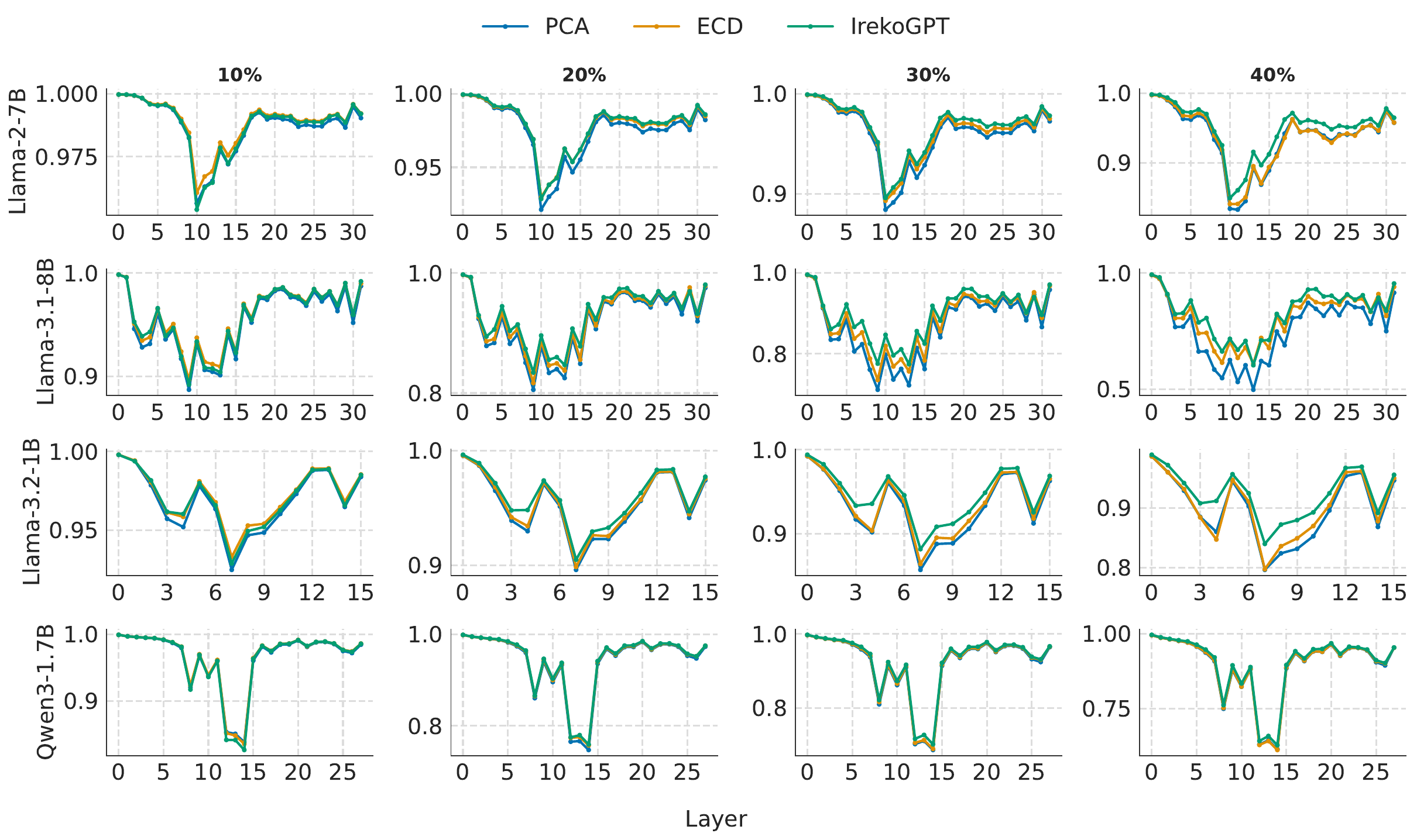}
    \caption{Layer-wise CKA between compressed and full-width value-projection outputs across
increasing sparsity levels. All models and all layers shown. We compare PCA-only, ECD, and \methname{}; higher is better.}
    \label{fig:all-models-cka}
\end{figure}
\begin{figure}[ht]
    \centering
    \includegraphics[width=0.99\linewidth]{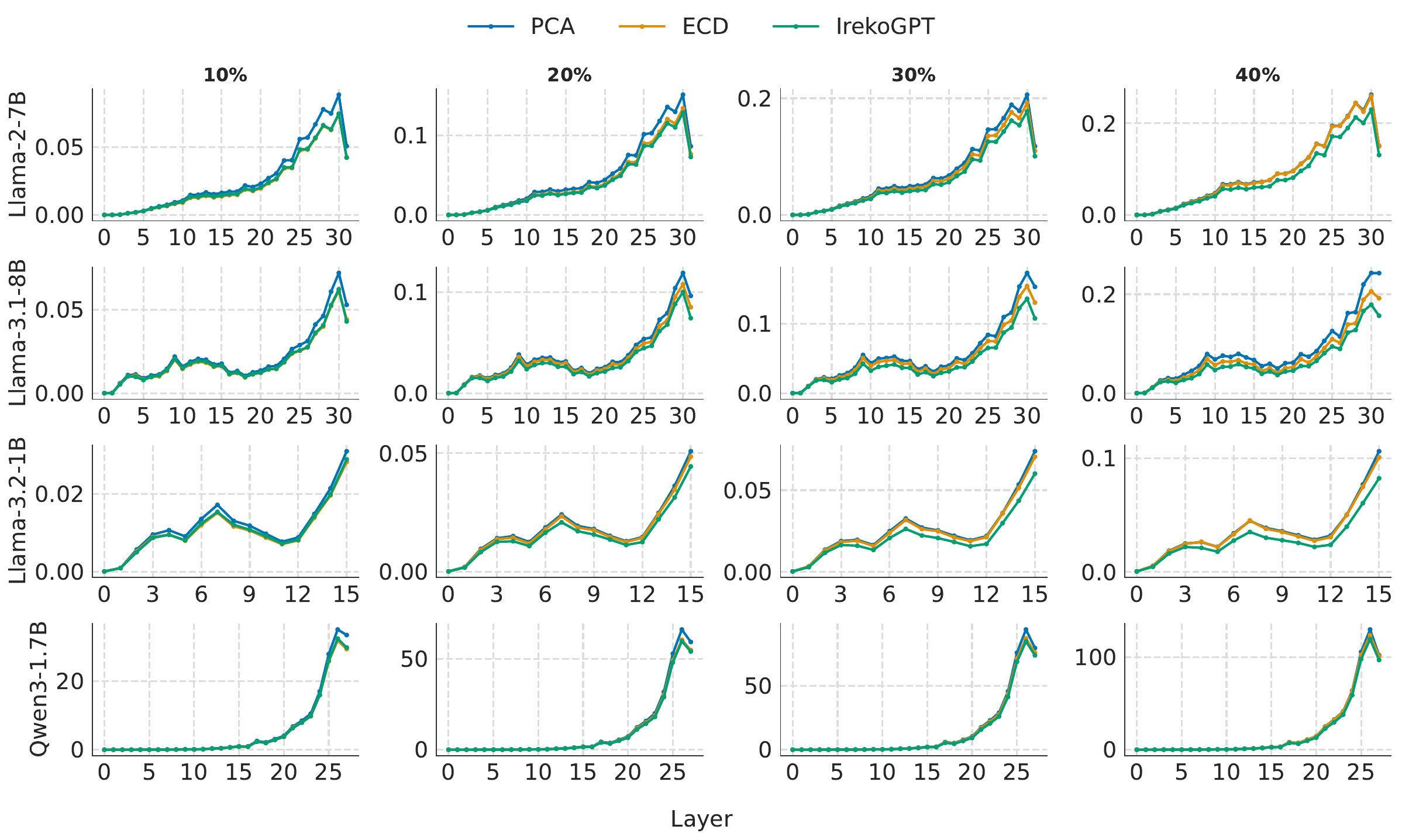}
    \caption{Layer-wise MSE between compressed and full-width value-projection outputs across
increasing sparsity levels. All models and all layers shown. We compare PCA-only, ECD, and \methname{}; lower is better.}
    \label{fig:all-models-mse}
\end{figure}
\begin{figure}[ht]
    \centering
    \includegraphics[width=0.99\linewidth]{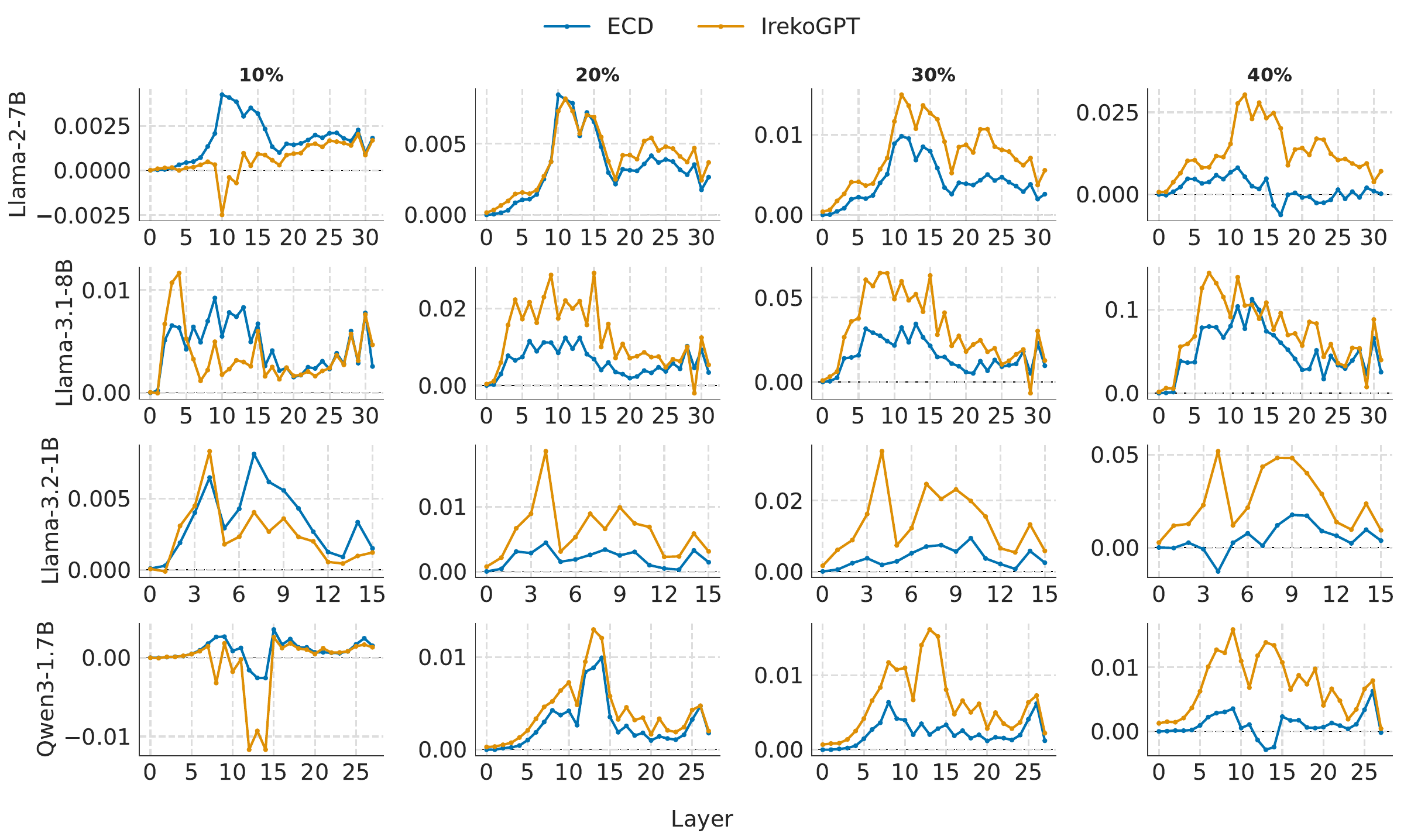}
    \caption{Layer-wise difference in CKA between ECD and \methname{} w.r.t PCA-only. CKA is still computed between compressed and full-width value-projection outputs.}
    \label{fig:all-models-cka-diff}
\end{figure}
\begin{figure}[ht]
    \centering
    \includegraphics[width=0.99\linewidth]{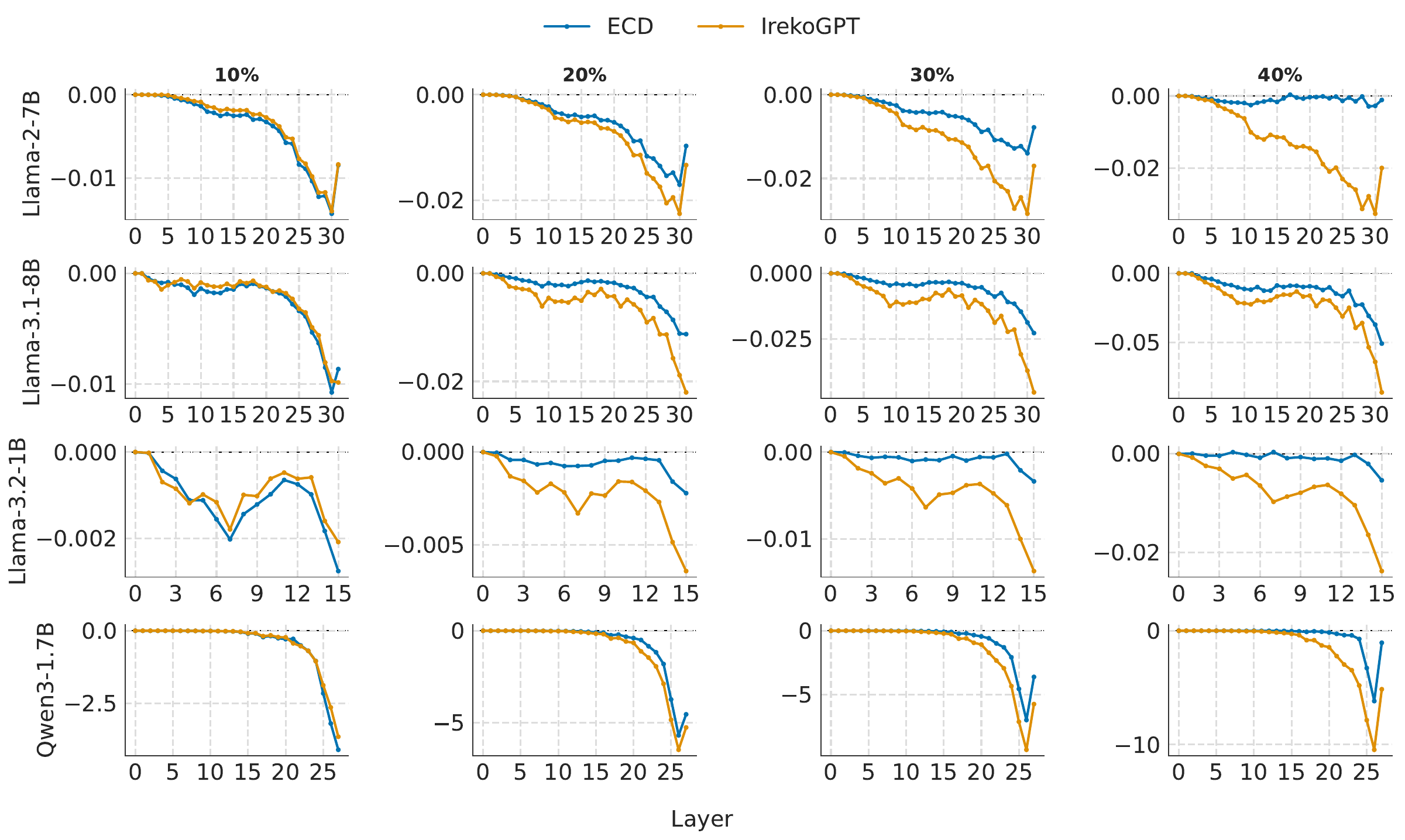}
    \caption{Layer-wise difference in MSE between ECD and \methname{} w.r.t PCA-only. MSE is still computed between compressed and full-width value-projection outputs.}
    \label{fig:all-models-mse-diff}
\end{figure}

\end{document}